\documentclass[11pt]{article}

\usepackage[final]{acl}
\usepackage{times}
\usepackage{latexsym}

\usepackage[T1]{fontenc}

\usepackage[utf8]{inputenc}

\usepackage{microtype}

\usepackage{inconsolata}

\usepackage{graphicx}
\usepackage{subcaption}
\usepackage{amsmath} 
\usepackage{amsfonts} 
\usepackage{booktabs} 
\usepackage{multirow} 
\usepackage{fontawesome5}
\newcommand{\cofirst}{\nobreak\hspace{0.15em}(\(\ddagger\))}
\newcommand{\corrauthor}{%
  \nobreak\hspace{0.15em}(\faIcon[regular]{envelope})%
}
\usepackage[table]{xcolor}
\title{AdaptInfer: Adaptive Token Pruning for Vision-Language Model Inference via Dynamical Text Guidance}

\author{
  \textbf{Weichen Zhang}\cofirst\textsuperscript{1,2},
  \textbf{Zhui Zhu}\cofirst\textsuperscript{2},
  \textbf{Ningbo Li}\textsuperscript{2,3},
  \textbf{Shilong Tao}\textsuperscript{4},\\
  \textbf{Hongzi Zhu}\textsuperscript{5},
  \textbf{Jingao Xu}\textsuperscript{1},
  \textbf{Kebin Liu}\corrauthor\textsuperscript{2},
  \textbf{Yunhao Liu\textsuperscript{2}}
\\
  \textsuperscript{1}The University of Hong Kong,
  \textsuperscript{2}Tsinghua University, \\
  \textsuperscript{3}The Hong Kong University of Science and Technology,
  \textsuperscript{4}Peking University, \\
 \textsuperscript{5}Shanghai Jiao Tong University
\\
weic.zhang@connect.hku.hk, z-zhu22@mails.tsinghua.edu.cn, nlibc@connect.ust.hk,\\ shilongtao@stu.pku.edu.cn,  hongzi@sjtu.edu.cn, jingaoxu@hku.hk,\\ \{kebinliu2021, yunhao\}@tsinghua.edu.cn\\
}

\begin{document}
\maketitle

\begingroup
\renewcommand{\thefootnote}{(\(\ddagger\))}
\footnotetext[0]{Equal contribution.}
\endgroup

\begingroup
\renewcommand{\thefootnote}{(\faIcon[regular]{envelope})}
\footnotetext[0]{Corresponding author.}
\endgroup

\begin{abstract}
 Vision–language models (VLMs) have achieved impressive performance on multimodal inference tasks, but the cost remains a significant challenge due to the large number of vision tokens processed during the prefill stage. Existing token pruning methods often rely on utilizing the static attention patterns directly, failing to exploit the dynamic internal signals within VLMs. To address the issue, we propose AdaptInfer, a novel plug-and-play framework for vision token pruning. First, we introduce a dynamic text-guided pruning mechanism that construct soft priors over text-token importance on each pruning layer, allowing more informed scoring of vision tokens at each stage. Second, we observe a highly consistent distribution of cross-modal attention shifts by architecture, which inspires us to introduce a efficient data-driven schedule which determines the pruning locations automatically. Experimental results have verified the effectiveness and generalization of the proposed method. Under the same token budget, AdaptInfer surpasses existing approaches in accuracy. For instance, AdaptInfer maintains averagely 99.4\% accuracy on Qwen2-VL while 70\% of the prefilling vision token overhead is reduced. The source code is available on: \url{https://github.com/weiczh02/AdaptInfer-base}.
\end{abstract}

\section{Introduction}

Vision--language models (VLMs) have achieved strong performance on multimodal tasks such as captioning~\citep{mscoco}, retrieval~\citep{vse}, and visual question answering~\citep{vqa}, by coupling visual encoders~\citep{convnet,vit} with large language model decoders~\citep{llama,gpt-3,vlm_survey}. Nevertheless, this architecture introduces a major inference bottleneck: visual inputs are represented by far more tokens than textual prompts. For example, a $672\times672$ image with a $14\times14$ patch size normally produces $2304$ visual tokens~\citep{clip,sparsevlm}, while the text prompt usually contains fewer than 100 tokens~\citep{gqa,mme}. Since vision tokens are often redundant and semantically repetitive~\citep{zhang2024VisPruner,flowcut}, reducing them has become a central direction for VLM acceleration.

Existing methods mainly prune, merge, or compress visual tokens either in the visual encoder~\citep{llama-vid,Recover} or during the LLM prefill stage~\citep{vtw,pdrop,fastv,mminference}. Recent attention-based methods further show that not all attention signals are equally useful for token ranking~\cite{abias}. VisPruner~\citep{zhang2024VisPruner} observes that full attention patterns can be noisy, while SparseVLM~\citep{sparsevlm} improves pruning by selecting salient text tokens as guidance before inference. Despite effectiveness, these methods still rely on static guidance. However, the informativeness of text tokens evolves as the model progressively refines its internal representations and activation patterns~\citep{bert-iner,whatdoesbertlookat,zhu2026dynamicinfer}. As shown in Figure~\ref{fig:miou}, the most important text tokens vary substantially across layers, suggesting that static text-token selection may lead to suboptimal pruning decisions and affect the performance.

Therefore, to truly harness the benefits of text-guided sparsification, it is crucial to develop a pruning strategy matched with dynamic fluidity of information, reflecting the effective cross-modal interaction throughout the inference process. Therefore, we propose to reconstruct the dynamic importance ranking of text tokens at each layer by utilizing the text-to-text (t2t) attention maps. These attentions provide a natural, layer-specific prior distribution for text-token importance, which we then use to reweight text-to-vision (t2v) attention scores for vision token ranking. Since these attention blocks are already available from the standard forward pass and are queried only at selected pruning layers, AdaptInfer introduces minor additional overhead.

Another challenge is how to choose pruning locations. Prior methods often rely on empirical rules or extensive hyperparameter search~\citep{pdrop,fastv,abias}, which can be computationally expensive and dataset-specific. We instead analyze layer-wise cross-modal attention shifts during inference and observe consistent shift patterns across datasets and architectures. For example, attention shifts concentrate around layers 1 and 10-20 in LLaVA-1.5-7B~\citep{liu2023llava}, and around layers 10-19 in Qwen2-VL-2B~\citep{Qwen2-VL}. These observations provide a simple, data-driven basis for selecting effective pruning layers without exhaustive tuning.

The solutions we propose effectively address the limitations of existing works in the field of vision token sparsification for VLM acceleration. Our main contribution involves:
\begin{itemize}
    \item We propose \textbf{AdaptInfer}, an adaptive vision token pruning method which dynamically determines text token guidance during inference. AdaptInfer is a plug-and-play solution. 
    \item We identify consistent cross-modal attention-shift patterns and use them to derive a simple, data-driven pruning schedule.
    \item We implement and evaluate our proposed solution, AdaptInfer, across multiple benchmarks and VLMs under different vision token budget settings. Within the same token budget, our AdaptInfer outperforms state-of-the-art (SOTA) methods on the inference accuracy.
\end{itemize}

\section{Related Work}
In this section, we will briefly introduce the previous works that are correlated with ours.
\subsection{Vision-Language Models}
Early multi-modal systems paired convolutional vision backbones with recurrent language decoders~\citep{caption_liff,showandtell}. Modern VLMs instead follow the Transformer paradigm~\citep{attention} by representing an image as a sequence of \emph{visual tokens} that interact with textual tokens in a shared self-attention space~\citep{liu2023llava,internvl1}. BLIP-2~\citep{blip2} and MiniGPT-4~\citep{minigpt4} introduce lightweight linear adapters that project features from a frozen CLIP encoder into the hidden space of a large language model, enabling efficient training. These explorations bridge the gap between frozen encoders and LLMs. The LLaVA family~\citep{liu2023improvedllava,liu2023llava,liu2024llavanext} refines this recipe with stronger instruction tuning, while a few other efforts such as Flamingo~\citep{flamingo} and CogVLM~\citep{cogvlm} scale the approach to billions of parameters.

Recent explorations also contribute to resolution and multi-modal extensions. Models leveraging hierarchical perception (e.g., LLaVA‑NeXT~\citep{liu2024llavanext}) and adaptive patching (e.g., Qwen‑VL~\citep{Qwen-VL}) permit high‑resolution inputs, while large-scale vision encoders are adopted to generate richer hidden states of the vision tokens~\citep{internvl1,internvl3}. However, these gains come at the cost of dramatically more vision tokens, which is a key bottleneck we target at.

\subsection{Inference Acceleration of VLMs}
Previous approaches mainly focus on vision token sparsification. This is because the number of vision tokens is often an order of magnitude (or more) larger than that of text tokens. In addition, visual embeddings are naturally much more sparse and repetitive than human-made texts~\citep{Marr2010}. In this field, there are two research directions including efficient vision encoders and vision token pruning in LLM networks. 

First, methods like LLaVA-PruMerge~\citep{prumerge}, VisionZip~\cite{visionzip}, and VisPruner~\citep{zhang2024VisPruner} follow the first direction, which cuts the encoder outputs or redesign the projector to produce fewer vision tokens. Recoverable compression~\citep{Recover} is also introduced to repair the information loss from token pruning within vision encoders. Solutions follow the second direction not only to drop the vision tokens~\citep{fastv,vtw,pdrop,mminference,multicue,fitprune} during prefill, but also merge them to compress the numbers of the vision tokens~\citep{tome}. SparseVLM~\citep{sparsevlm} tries to go deeper in exploration of static text prompt guidance but ignoring the evolving inherent of token information. Our approach contributes to the second paradigm.

More broadly, structured representations have been explored as an effective way of capturing complex relationships and dependencies in learning-based systems~\citep{tao2025unisoma,tao2026neural}. Recent studies have further leveraged such explicit structural modeling, like graph-structured representations, to guide visual token pruning in multimodal models~\citep{g-prune,yang2025script}.

\newcommand{\hl}[1]{\cellcolor{blue!15}#1}
\section{Method}

In this section, we will introduce our observations and the methods we propose.

\begin{figure}[t]
    \centering
    \includegraphics[width=0.85\linewidth]{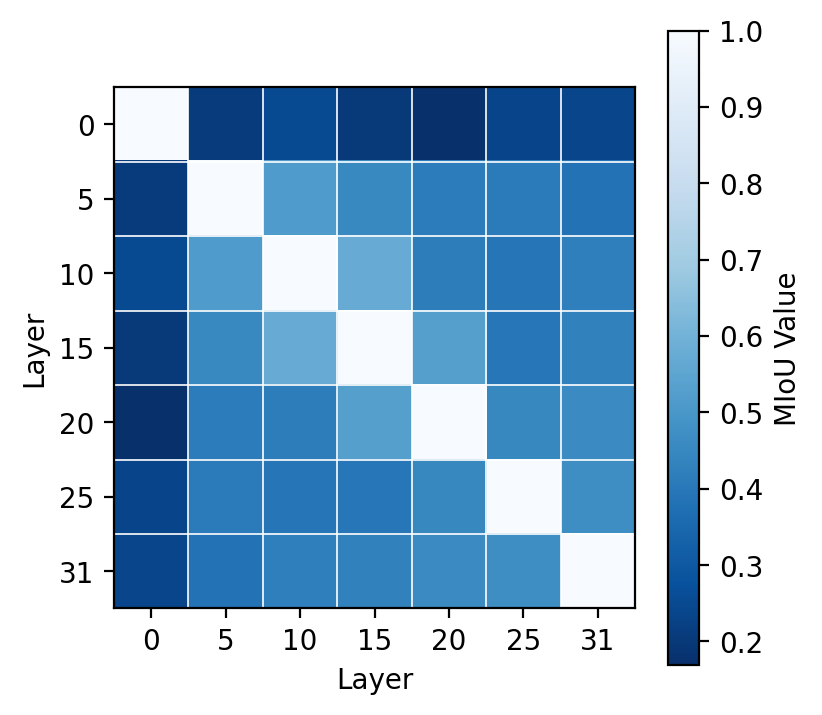}
    \caption{\textbf{MIoU of key text tokens in different layers.} The consistent low mIoU shows the key text token varies across layers.}
    \label{fig:miou}
\end{figure}

\begin{figure}[t]
    \centering
    \includegraphics[width=0.9\linewidth]{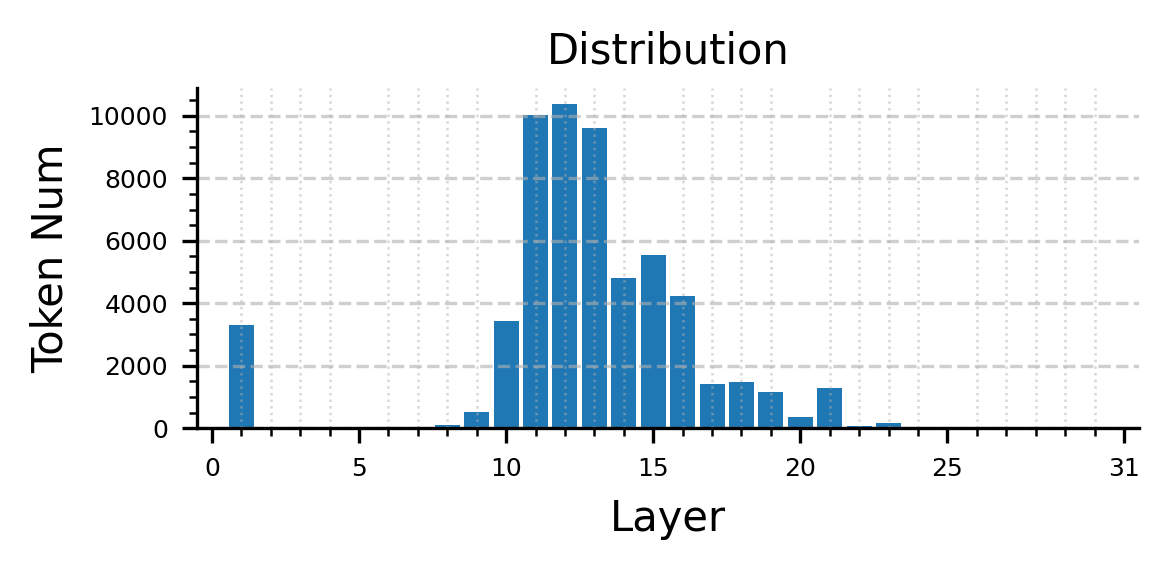}\\
    \small (a) Attention shift distribution on MME.\\

    \includegraphics[width=0.9\linewidth]{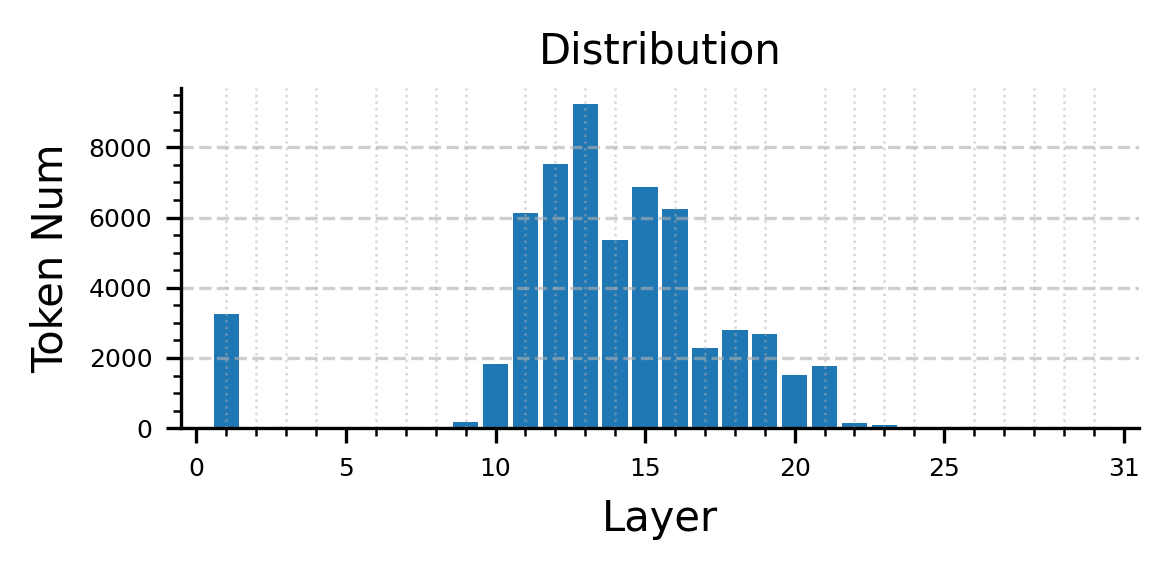}\\
    \small (b) Attention shift distribution on TextVQA.
    
    \caption{\textbf{Highly consistent distribution of attention shifts on MME and TextVQA.}}
    \label{fig:attn_distribution}
  
\end{figure}
\subsection{Observations}
We present a few observations and preliminary experiments that have illuminated our insights.

\subsubsection{Text Token Importance}
SparseVLM~\citep{sparsevlm} selects key text token offline for pruning guidance, which actually based on an assumption that the text token importance remains static across all prefilling layers. Since this assumption is not trivial, we examine it as follows.

Given an input question prompt, we first tokenize it as \(Prompt=[t_1,\ldots,t_n]\). We then define the \textit{importance} of a text token \(t_i\) at layer \(\ell\) as the total attention weight  it receives from all text tokens in the layer-\(\ell\) text-to-text (t2t) attention map on average of attention heads:
\begin{equation}
  \mathrm{Imp}_\ell(t_i)= \frac{1}{H}\sum_{h=1}^{H}\sum_{j=1}^{n} \mathbf{A}^{(\ell,h)}_{\text{t2t}} [j,i],
\end{equation}
where \(\mathbf{A}^{(\ell,h)}_{\text{t2t}} [j,i]\) denotes the attention of head $h$ directed from token \(t_j\) to \(t_i\), $H$ represents the number of the attention heads. Intuitively, tokens with high
\(\mathrm{Imp}_\ell\) are the current key text tokens of
the language stream and are therefore best suited to guide cross-modal
pruning decisions at that layer.

\begin{figure*}[t]
    \centering
    \includegraphics[width=1\linewidth]{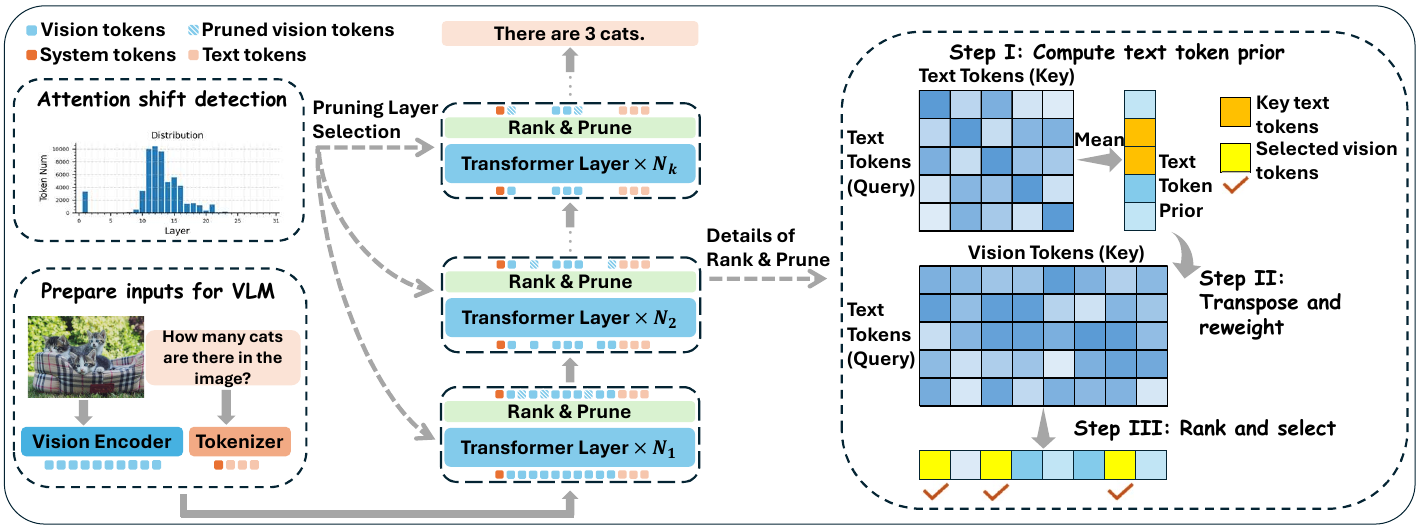}
    \caption{\textbf{The architecture of AdaptInfer}. Text token importance is computed and guides vision token selection at every pruning layer adaptively. The right panel illustrates the implementation details of the Rank \& Prune module. }
    \label{fig:structure}
\end{figure*}

\subsubsection{Dynamics of Text Token Importance}

To examine whether text-token importance remains stable across layers, we conduct an empirical study on LLaVA-1.5-7B~\citep{liu2023llava} using 1,000 samples from TextVQA. For each selected layer, we compute text-token importance from the text-to-text attention map and select the top 20\% tokens as key text tokens. We then measure the mean Intersection over Union (mIoU) of these key-token sets across layers, as shown in Figure~\ref{fig:miou}.

The consistently low off-diagonal mIoU values indicate that key text tokens vary substantially during inference. For example, the mIoU between layer 0 and layer 24 is only 0.169, meaning that most selected key tokens differ across these two layers. This suggests that VLMs attend to different parts of the input question at different reasoning stages. Consequently, any static text-prompt–guided pruning approach is likely to fail in capturing this evolving semantic alignment. These findings motivate our dynamical, layer-wise text-guidance mechanism.

\subsubsection{Cross-Attention Shifts of VLM}\label{sec:changepoint}

To derive pruning schedules from the model's own inference dynamics rather than from trial-and-error tuning, we analyze where cross-modal attention shifts occur across layers. Specifically, for each sample in LLaVA-1.5-7B~\citep{liu2023llava}, we compute cumulative text-to-vision (t2v) attention scores for visual tokens and focus on the top 10\% vision tokens receiving the highest total attention in the prefill stage. These tokens are treated as the most task-relevant visual evidence for the corresponding multimodal input.

We then apply change-point detection~\citep{changpoint} to the layer-wise cumulative attention trajectory of each selected token, identifying layers where its attention pattern changes significantly. Appendix~\ref{appendix:cpd} shows the technical details. Such shift points indicate layer-wise transitions in visual information usage, either when a token becomes salient or when its information has largely been consumed.

Figure~\ref{fig:attn_distribution} aggregates the detected shift locations over 1,000 samples from both MME~\citep{mme} and TextVQA~\citep{textvqa}. Despite their different task formulations and data distributions, the two datasets exhibit highly consistent patterns: attention shifts concentrate around layer 1 and layers 10--20, while layers 2--9 and later layers are attention stable regions. This consistency suggests that the observed attention dynamics are more likely to reflect architecture-level behavior than dataset-specific artifacts. Therefore, these shift distributions provide a data-driven basis for selecting pruning layers according to stable attention regions, rather than relying solely on empirical heuristics.

\subsection{Adaptive Token Pruning}

Based on the two key observations above, we propose AdaptInfer for VLM inference acceleration. 

\subsubsection{Dynamic Text Guidance Pruning}

To guide visual token pruning in a more informed and adaptive way, we propose a dynamic text-guidance mechanism that leverages the model’s internal attention signals. Instead of statically selecting a fixed subset of text tokens before entering the language model~\citep{sparsevlm}, we dynamically infer their relative importance during inference at each predefined pruning layer. We reuse the attention maps already computed by the model. The architecture of Our AdaptInfer is shown in Fig~\ref{fig:structure}, and our dynamic text guidance mechanism can be divided into three steps.

Firstly, in each pruning layer, we extract the \textit{text-to-text} (t2t) attention matrix $\mathbf{A}^h_{\text{t2t}} \in \mathbb{R}^{T \times T}$, where $T$ is the number of text tokens, $h$ is the index of attention head with a total number of $H$. To estimate the importance of each text token, we aggregate attention scores across the query dimension:

\begin{equation}\label{eqa:2}
    \mathbf{w^{(h)}} =  \sum_{i=1}^{T} 
    \mathbf{A}^{(h)}_{\text{t2t}}[i, :] 
    \in \mathbb{R}^{T},
\end{equation}

where $\mathbf{w}$ serves as a soft prior distribution over the text tokens, averaged across all attention heads, indicating how much attention each token receives from the rest of the sequence.

Then, we reweight the t2v attention matrix $\mathbf{A}^{(h)}_{\text{t2v}} \in \mathbb{R}^{T \times V}$ with this prior, where $V$ is the number of visual tokens remained. $\mathbf{A}^{(h)}_{\text{t2v}}$ denotes the cross-attention matrix where text tokens are used as queries, and vision tokens serve as keys and values. The importance score of each visual token on average of all attention heads is computed as:

\begin{equation}\label{eqa:3}
    \mathbf{s} = \frac{1}{H}\sum_{h=1}^{H} \mathbf{w}^{(h)\top} \cdot \mathbf{A}^{(h)}_{\text{t2v}} \in \mathbb{R}^{V}.
\end{equation}

Here, $\mathbf{s}_j$ reflects the aggregated and weighted attention from all text tokens to visual token $j$. 

Finally, based on these scores, we rank all visual tokens and retain the top-$k$ for the current layer:

\begin{equation}
    \mathcal{I}_k = \text{TopK}(\mathbf{s}, k).
\end{equation}

Importantly, all text tokens participate in visual token scoring, but contribute in proportion to their dynamically inferred importance. Moreover, because both $\mathbf{A}_{\text{t2t}}$ and $\mathbf{A}_{\text{t2v}}$ are natively computed in standard forward passes, our method introduces little additional computational overhead. Note that this solution follows a training-free paradigm and can be seamlessly integrated as a plugin into existing VLMs. The detailed FLOPs derivations are also provided in Appendix~\ref{sec:compute}.

\subsubsection{Heuristic Pruning Schedule}

Following the attention-shift analysis described above, we design a pruning schedule that aligns with the layer-wise evolution of cross-modal attention. In our implementation, pruning at layer $\ell$ indicates to perform token pruning after the forward computation of layer $\ell$: we use the attention scores produced at this layer to rank vision tokens, remove low-ranked tokens, and feed only the remaining tokens into layer $\ell+1$. Therefore, a pruning location should be interpreted as the boundary after layer $\ell$.

Prior works~\cite{fastv, pdrop, sparsevlm} directly rely on attention scores to rank tokens and select pruning locations manually. However, we argue that attention scores from different transformer layers are not equally reliable as pruning evidence. Our attention-shift analysis reveals two typical regimes. In high-variation regions, token importance is actively being reassigned, with tokens either becoming newly important or having already been exploited by the model. Attention rankings in these regions are unstable, making pruning decisions less reliable. In stable regions, by contrast, token importance rankings remain relatively consistent and therefore provide more reliable pruning signals.

\begin{table*}[ht]
\small
\centering

\caption{\textbf{Comparison of methods under different pruning budgets on LLava-1.5-7B}.}
\label{tab:ours_pruning_compare}
\begin{tabular}{l|c|cccccc|c}
\toprule

\textbf{Method} & \textbf{Avg. Tokens} & \textbf{MME} &\textbf{GQA} & \textbf{MMB} & \textbf{SQA} &\textbf{TextVQA}&\textbf{MMVet} &\textbf{Ratio (\%)}  \\
\midrule
Vanilla  & 576        &1864  & 61.9 & 64.6   & 69.5 & 58.3 & 31.1& 100  \\
\midrule
ToMe (ICLR 23)   & 128 & 1343 & 52.4 & 53.3   & 59.6 & 49.1 & 27.2  & 82.8 ($\downarrow$ 17.2)  \\
FastV (ECCV 24)  & 128 & 1490 & 49.6 & 56.1   & 68.6 & 52.5 & 26.3  &87.9 ($\downarrow$ 12.1) \\
PDrop (CVPR 25)  & 128 & 1761 & 57.1 & 61.6   & 68.4 & 56.6 &  28.1 &94.7 ($\downarrow$ 5.3) \\
SparseVLM (ICML 25) & 128 & 1746& 58.4 & 64.5 & 68.6 & 56.7 & 29.0  &96.2 ($\downarrow$ 3.8) \\
DART (EMNLP 25) & 128& 1804 & 57.7 & 62.3 &69.4 &55.9 & 29.8 & 96.3 ($\downarrow$ 3.7)\\
A-Bias (arXiv 26.1)& 128 &1771 & 58.8 & 63.8 & 68.0 & 56.6  &29.2 & 96.3 ($\downarrow$ 3.7)\\
\rowcolor{blue!12}
AdaptInfer (Ours)   & 128 & 1794 & 58.5 & 63.8 & 69.9 & 56.8  & 29.5 & 97.1 \color{gray}{($\downarrow$ 2.9)}  \\
\midrule
ToMe (ICLR 23)   & 64  & 1138 & 48.6 & 43.7  & 50.0 & 45.3 & 24.1& 72.4 ($\downarrow$ 27.6) \\
FastV (ECCV 24)  & 64  & 1255 & 46.1 & 47.2  & 68.7 & 45.9 & 19.6 & 75.9 ($\downarrow$ 24.1)\\
PDrop (CVPR 25)  & 64 & 1561 & 47.5 & 58.8  & 69.0 & 50.6 & 22.5 & 85.0 ($\downarrow$ 15.0)\\
SparseVLM (ICML 25)    & 64 & 1589 & 53.8 & 60.1 & 69.8 & 53.4 & 24.9 & 89.5 ($\downarrow$ 10.5)\\
DART (EMNLP 25) & 64& 1533& 51.2 &55.8 &68.7 &50.4& 25.1 & 86.2 ($\downarrow$13.8)\\
A-Bias (arXiv 26.2)& 64 &1623 & 54.0 & 59.7 &  69.2 &  53.4 & 25.9 & 90.1 ($\downarrow$ 9.9)\\
\rowcolor{blue!12}
AdaptInfer (Ours)   & 64 & 1684 & 53.2 & 61.7  & 69.9 & 54.3 & 25.9& 91.5 \color{gray}{($\downarrow$ 8.5)} \\
\bottomrule
\end{tabular}
\end{table*}

Formally, let the model have $L$ transformer layers and let $K$ denote the default number of pruning stages. From the offline attention-shift analysis, we detect the starting layers of stable regions:
\begin{equation}
\begin{aligned}
\mathcal{S} &= \{s_1,s_2,\ldots,s_N\}, \\
0 &\leq s_1 < s_2 < \cdots < s_N < L ,
\end{aligned}
\end{equation}
where $N$ is the number of detected stable-region starts. Since pruning at layer $\ell$ removes tokens before layer $\ell+1$, a stable region starting at $s_i$ should be associated with the boundary right before it, i.e., after layer $s_i-1$. Pruning at this boundary allows the entire following stable region to operate on fewer visual tokens, thus maximizing computation savings while still relying on a stable attention-ranking regime. We therefore convert the detected stable-region starts into a set of pruning layer candidates:
\begin{equation}
\tilde{\mathcal{S}}
=
\left\{
\tilde{s}_i \mid \tilde{s}_i=\max(s_i-1,0),\ i=1,\ldots,N
\right\}.
\end{equation}

If $N \geq K$, we choose the first $K$ candidates as the pruning layers:
\begin{equation}
\mathcal{P}=\{\tilde{s}_1,\tilde{s}_2,\ldots,\tilde{s}_K\}.
\end{equation}
If $N<K$, we keep all detected candidates and fill the remaining $K-N$ pruning layers by uniform interpolation between the earliest two pruning candidates. This interpolation aims to distribute pruning across depth. It places additional pruning stages early to maximize computation savings, while maintaining sufficient layer intervals for each pruning step to take effect. Notably, the interpolation strategy is an efficiency-oriented scheduling heuristic rather than another stable guarantee. Specifically, the additional pruning layers are
\begin{equation}
\begin{aligned}
\mathcal{U}
= \Big\{
&\left\lfloor
\tilde{s}_1+\frac{j}{K-N+1}(\tilde{s}_2-\tilde{s}_1)
\right\rfloor \\
&\mid\ j=1,2,\ldots,K-N
\Big\}.
\end{aligned}
\end{equation}
The final pruning schedule is then
\begin{equation}
\mathcal{P}=\operatorname{Sort}(\tilde{\mathcal{S}}\cup\mathcal{U}).
\end{equation}

We set $K=3$ in our experiments, the ablation study on $K$ is included in Table~\ref{tab:ablation2}. In our experiments, the attention-shift analysis typically identifies two stable regions on Qwen, LLaVA, and InternVL models. Thus, under the default setting $K=3$, the schedule reduces to selecting the two stable-region boundaries and their midpoint. For example, on LLaVA-1.5-7B, the detected stable starts correspond to pruning candidates layers $1$ and $20$, yielding the schedule $\{1,10,20\}$. For LLaVA-1.5-13B, we use $\{1,11,22\}$; for Qwen2-VL-2B~\citep{Qwen2-VL}, we use $\{0,9,19\}$. Additional experiments on LLaVA-1.5-13B and Qwen2-VL-2B are presented in Appendix~\ref{13b} and Appendix~\ref{qwen2b}. The proposed offline analysis is lightweight: it is performed only once for each VLM and typically finishes within minutes.

\subsubsection{Discussion}

AdaptInfer only accesses attention at the selected pruning layers, at most three in our experiments, and uses only the \textit{text-to-text} (\(t2t\)) and \textit{text-to-vision} (\(t2v\)) blocks for token ranking, without materializing full attention maps or the much larger \textit{vision-to-vision} (\(v2v\)) block. This can be implemented with the FlashAttention-compatible dual-pass strategy used in SparseVLM~\cite{sparsevlm}. Since attention is queried only at a few pruning layers and only for the required sub-blocks, the resulting end-to-end latency overhead is modest in practice. We defer the implementation details and complexity discussion to Appendix~\ref{sec:flashattention}.

\section{Experiment}

\begin{table*}[ht]
\small
\centering
\caption{\textbf{Comparison of Performance on Qwen2-VL-2B.}}
\label{qwen-exp}
\begin{tabular}{l|c|ccccc|ccc|c}
\toprule
\multirow{2}{*}{\textbf{Methods}} & \multirow{2}{*}{\textbf{Tokens}} & 
\multicolumn{5}{c|}{\textbf{Image-Based}} & 
\multicolumn{3}{c|}{\textbf{Video-Based}} & 
\multirow{2}{*}{\textbf{Ratio}} \\
& & \textbf{MME} & \textbf{TextVQA} & \textbf{GQA} & \textbf{MMB} & \textbf{POPE} & 
\textbf{TGIF} & \textbf{MSRVTT} & \textbf{MSVD} &  \\
\midrule
Vanilla    & 100\% & 1901 & 77.8 & 60.4 & 71.7 & 86.9 & 9.9  & 30.9 & 42.3 & 100\% \\
\midrule
SparseVLM       & 50\%  & 1900 & 76.9 & 59.9 & 71.0 & 86.4 & 11.7 & 31.2 & 42.9 & 102.1\% \\
\rowcolor{blue!12}
AdaptInfer      & 50\%  & 1909 & 76.6 & 60.0 & 72.1 & 86.7 & 11.1 & 30.5 & 43.6 & 101.7\% \\
\midrule
SparseVLM       & 30\%  & 1867 & 73.6 & 57.5 & 70.1 & 84.6 & 9.3  & 29.5 & 41.8 & 96.4\% \\
\rowcolor{blue!12}
AdaptInfer      & 30\%  & 1885 & 73.9 & 58.5 & 71.4 & 85.8 & 10.4 & 30.2 & 43.5 & 99.4\% \\
\midrule
SparseVLM       & 10\%  & 1460 & 51.3 & 40.4 & 57.3 & 40.3 & 5.0  & 28.1 & 30.2 & 68.6\% \\
\rowcolor{blue!12}
AdaptInfer      & 10\%  & 1721 & 62.3 & 48.8 & 63.7 & 71.3 & 5.9  & 28.2 & 41.4 & 81.6\% \\
\bottomrule
\end{tabular}
\end{table*}

In this section, we empirically evaluate our proposed approach through a series of experiments.
\subsection{Experimental Settings}

\subsubsection{Datasets}
For multimodal evaluation, we test our solutions on several widely used benchmarks, including MME~\citep{mme}, GQA~\citep{gqa}, MMBench (MMB)~\citep{mmbench}, ScienceQA (SQA)~\citep{sqa}, TextVQA~\citep{textvqa}, MMVet~\citep{mmvet}, and POPE~\citep{pope}. These datasets together provide a comprehensive evaluation, including vision-question-answering (VQA), optical character recognition (OCR), perception, reasoning, factual grounding and so on.
In addition, we also test our solutions on three popular video-based multimodal benchmarks, TGIF-QA (TGIF)~\citep{tgif}, MSVD-QA (MSVD) and MSRVTT-QA (MSRVTT)~\citep{MSRVTTQA}.

\subsubsection{Baselines}
We select six representative vision-token sparsification methods that operate in a plug-and-play manner during the LLM forward pass, including FastV~\citep{fastv}, ToMe~\citep{tome}, Pyramid Drop (PDrop)~\citep{pdrop}, SparseVLM~\citep{sparsevlm}, DART~\citep{dart}, and Attention-Bias (A-Bias)~\citep{abias}. For PDrop, we adopt its training-free version to ensure a fair comparison under the same deployment setting. For Attention-Bias, we use its PDrop-based variant for evaluation.

We focus the main comparison on methods in which visual token sparsification happens in the language model prefill stage in Section~\ref{main-results}. Methods that reduce visual tokens at different locations, such as vision-encoder-side compression~\citep{visionzip}, are orthogonal to our prefill-stage pruning. To verify this complementarity, we further combine VisionZip with AdaptInfer and report the results in Section~\ref{vision-encoder-compression}.

\subsubsection{Implement Details}
To ensure a comprehensive comparison between baselines and AdaptInfer, we report the results in different average retained token budgets. AdaptInfer utilizes the pruning parameters described above while the baselines keep their original settings for all core components. Our experiments covers three different types of VLMs, including LLaVA-1.5-7B~\cite{liu2023llava}, Qwen2-VL-2B~\cite{Qwen2-VL}, and InternVL3-2B~\cite{internvl3}. Additional experiments on LLaVA-1.5-13B are presented in appendix~\ref{13b-exp}.

\subsection{Main Results on LLaVA}\label{main-results}

\begin{figure}[t]
    \centering
    \begin{subfigure}[t]{\linewidth}
    \centering
    \includegraphics[width=0.8\linewidth]{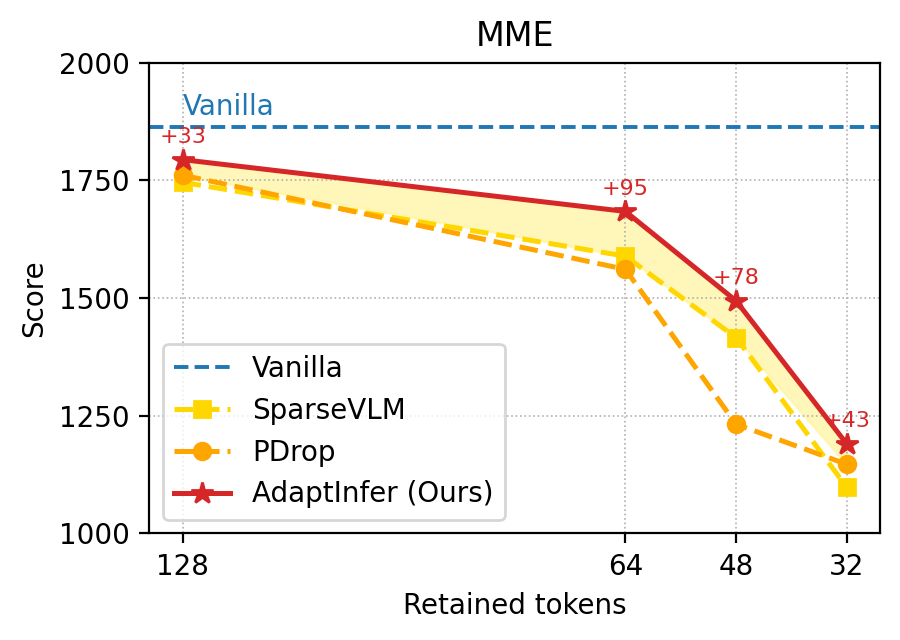}\\
    \small (a) Performance trends on MME.
    \end{subfigure}
    \hfill
    \begin{subfigure}[t]{\linewidth}
    \centering
    \includegraphics[width=0.8\linewidth]{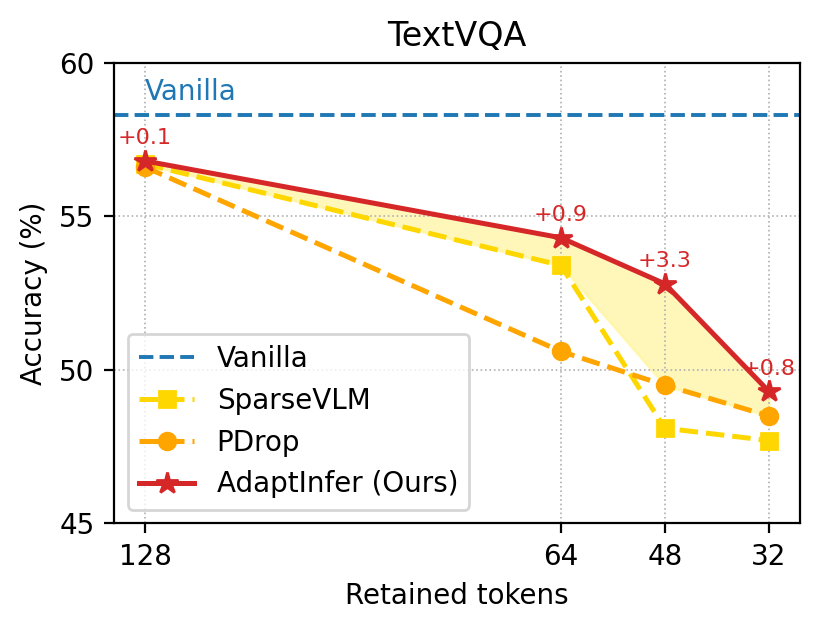}\\
    \small (b) Performance trends on TextVQA.
    \end{subfigure}

    \caption{\textbf{Performance trends of AdaptInfer on two representative benchmarks.} }
    \label{fig:trend}
\end{figure}

\begin{table}[t]
\small\setlength{\tabcolsep}{3pt}
\centering
\caption{\textbf{Extreme pruning comparsions.}}
\label{tab:ablation1}
\begin{tabular}{l|c|cc|c}
\toprule
\textbf{Method} & \textbf{Tokens} & \textbf{MME} &\textbf{TextVQA}& \textbf{Ratio (\%)} \\
\midrule
Vanilla  & 576        &1864& 58.3 & 100 \\
\midrule
SparseVLM& 48 & 1416  & 48.1 & 79.2   \\
\rowcolor{blue!12}
AdaptInfer   & 48 & 1494 & 52.8 & 85.4 \\

\midrule
SparseVLM & 32 & 1098 & 47.7 & 70.4 \\
\rowcolor{blue!12}
AdaptInfer   & 32 & 1190 & 49.3 & 74.2 \\

\bottomrule
\end{tabular}
\end{table}

Table~\ref{tab:ours_pruning_compare} reports the results on LLaVA-1.5-7B under average vision-token budgets of 128 and 64. AdaptInfer achieves the best overall performance in both settings, retaining 97.1\% and 91.5\% of the full-token accuracy, respectively. Notably, with only 64 retained tokens, it reduces the prefill vision-token load by 88.9\% while outperforming the strongest baselines SparseVLM and A-Bias by 2.0\% and 1.4\%. Although the absolute gains appear modest, they are achieved in a near-saturated regime where all strong baselines are already close to the full-token upper bound. Moreover, we shows the consistent gains of AdaptInfer over SparseVLM and PDrop across token budgets from 128 to 32 in figure~\ref{fig:trend}.



\begin{table*}[ht]
\centering
\small
\caption{\textbf{Complementarity with vision-encoder-side compression.} }
\begin{tabular}{l|c|ccccc|c}
\toprule
\textbf{Method} & \textbf{Tokens} & \textbf{TVQA} & \textbf{MME} & \textbf{MMB} & \textbf{POPE} & \textbf{GQA} & \textbf{Ratio}\\
\midrule
Vanilla & 576 & 58.3 & 1864 & 64.6 & 85.9 & 61.9 &100\%\\
\midrule
VisionZip & 64 & 55.5 & 1690 & 60.1 & 77.0 & 55.1  & 91.5\%\\
\rowcolor{blue!12}
VisionZip + AdaptInfer & 64 & 56.5 & 1724 & 62.8 & 83.3 & 56.6 &95.0\% \\
\bottomrule
\end{tabular}

\label{tab:visionzip_adaptinfer}
\end{table*}

\begin{table*}[ht]
\centering
\small
\caption{\textbf{Latency test of AdaptInfer.}}
\label{tab:Latency}
\begin{tabular}{l|c|c|cccc|c}
\toprule
\textbf{Method} & \textbf{Tokens} &\textbf{Metrics} & \textbf{MME} & \textbf{GQA} & \textbf{MMB}  & \textbf{TextVQA} & \textbf{Average} \\
\midrule
Vanilla        & 576 & FLOPs (T) & 4.268 & 4.250 & 4.623 & 4.611 & 4.438 \\
                &     & Latency (ms) & 82.0 & 76.5 & 91.3 & 91.2 & 85.3 \\
\midrule
PDrop     & 64  & FLOPs (T)& 0.975 & 0.958 & 1.316  & 1.305 &1.138 ($\downarrow$ 74.3\%)\\
                &     & Latency (ms) & 33.0 & 32.0 &36.4  & 36.6  & 34.5 ($\downarrow$ 59.6\%)\\
                
\midrule
SparseVLM       & 64 & FLOPs (T)& 0.974 & 0.958 & 1.316  & 1.305 &1.138 ($\downarrow$ 74.3\%)\\
                &     & Latency (ms) & 34.4 & 34.7 &38.1  & 39.5  & 36.7 ($\downarrow$ 57.0\%)\\
\midrule
\rowcolor{blue!12}
AdaptInfer (Ours) &  64  & FLOPs (T)& 0.975 & 0.959 & 1.317  & 1.305 & 1.139 ($\downarrow$ 74.3\%) \\

\rowcolor{blue!12}
 \quad &   &  Latency (ms) & {31.0} & {30.9} & {34.5} & {35.5} & {33.0 ($\downarrow$ 61.3\%)} \\
\bottomrule
\end{tabular}
\end{table*}

We further evaluate AdaptInfer on a more recent VLM Qwen2-VL~\citep{Qwen2-VL} to examine its generalization across different VLM architectures. For Qwen2-VL, which dynamically adjusts the number of visual tokens and supports both image and video tasks, we compare AdaptInfer with SparseVLM under the same retained-token ratios of 50\%, 30\%, and 10\%. As shown in Table~\ref{qwen-exp}, AdaptInfer consistently matches or outperforms SparseVLM across eight benchmarks, including three video QA datasets TGIF, MSRVTT, and MSVD. Notably, at the 50\% retained ratio, both pruning methods slightly outperform the full-token model, suggesting that removing redundant visual tokens can reduce noise while improving efficiency. For more implementation details of AdaptInfer on Qwen2-VL, please refer to Appendix~\ref{appendix:qwen_detail}.

We also examine our method under extreme pruning budgets on LLaVA-1.5-7B, which is presented in Table~\ref{tab:ablation1}. AdaptInfer still has considerable performance and consistently outperforms SparseVLM when retaining only 48 or 32 visual tokens on average, preserving 85.4\% and 74.2\% of the full-token accuracy respectively.

\subsection{Complementarity with Vision-Encoder Compression}\label{vision-encoder-compression}

AdaptInfer performs token pruning during the LLM prefill stage and is therefore orthogonal to methods that compress visual tokens before they enter the language model. To verify this complementarity, we combine AdaptInfer with VisionZip~\citep{visionzip}, a vision-encoder-side token compression method, and evaluate whether dynamic text-guided pruning remains effective after upstream visual token reduction. As shown in Table~\ref{tab:visionzip_adaptinfer}, adding AdaptInfer on top of VisionZip consistently improves performance across all evaluated benchmarks under the same average token budget of 64.

\subsection{Efficiency Analysis}

To approximate a resource-constrained deployment setting, we evaluate the practical efficiency of AdaptInfer on LLaVA-1.5-7B using CUDA latency measurements on a single NVIDIA RTX 4090 GPU with FLOPs estimation, as reported in Table~\ref{tab:Latency}. To be specific, we evaluate the same method-specific 64-token settings here whose accuracy results are reported in Table~\ref{main-results}. All latency results are averaged per sample. Under the same 64-token budget, AdaptInfer achieves an average latency of 33.0 ms, lower than both PDrop~\citep{pdrop} and SparseVLM~\citep{sparsevlm}. Although the theoretical FLOPs of these methods are comparable, AdaptInfer attains lower real latency because it introduces little additional computation, requiring neither extra attention computation nor offline feature matching.

\subsection{Ablation Study}

\begin{table}[t]
\small\setlength{\tabcolsep}{4pt}
\centering
\caption{\textbf{Controlled ablation of different text priors on InternVL3-2B.} }
\label{tab:ablation_prior}
\begin{tabular}{l|c|cccc}
        \toprule
        \textbf{Tokens} & \textbf{Text Prior} & \textbf{MME} &\textbf{TextVQA} &\textbf{SQA} &\textbf{POPE}\\
        \midrule
        
        30\% & Single-token      & 1878    & 35.9 & 86.3 &84.7 \\
        30\% & Random            & 1990  & 48.6  & 88.6 & 86.4          \\
        30\% & Static             & 1989 & 49.3&88.7&86.9  \\
        \rowcolor{blue!12}
        30\% & Dynamic &2015 &49.4  & 89.5   & 87.3  \\
        \bottomrule
    \end{tabular}
\end{table}

\begin{table}[t]
\small
\centering
\caption{\textbf{Performance of different pruning locations on LLava-1.5-7B.} }
\label{tab:ablation2}
\begin{tabular}{l|c|c|cc}
        \toprule
        \textbf{Tokens} & \textbf{K} & \textbf{Pruning Layers}  & \textbf{MME} & \textbf{MMB} \\
        \midrule
        \rowcolor{blue!12}
        64&3      & 1,10,20    & 1684 & 61.7 \\
        \midrule
        64&2              & 1,20      & 1520          & 56.3          \\
        64&4              & 1,7,14,20 & 1645          & 60.7          \\
        \midrule
        64&3              & 0,10,20   & 1660          & 60.7          \\
        64&3              & 2,10,20   & 1589          & 60.1          \\
       64& 3              & 1,9,20    & 1642          & 60.9          \\
       64& 3              & 1,11,20   & 1656          & 61.7          \\
       64& 3              & 1,10,19   & 1673          & 61.9          \\
       64& 3              & 1,10,21   & 1675          & 61.5          \\
        \bottomrule
    \end{tabular}
    \vspace{-1em}
\end{table}

We first evaluate the necessity of dynamic text guidance through a controlled ablation study on InternVL3-2B~\cite{internvl3}. We keep the pruning layers, token budget, per-stage pruning ratios, implementation, and evaluation protocol identical, and change only the text-query weighting strategy. The results are shown in Table~\ref{tab:ablation_prior}. In the table, \textit{Random prior} indicates a prior sampled from a symmetric Dirichlet distribution (five runs averaged); \textit{single-token prior} refers to a one-hot prior that assigns all weight to the first text token after the visual-token sequence; \textit{static prior} is a SparseVLM-style fixed prior that remains unchanged across pruning stages, without the clustering-and-merging component of the full SparseVLM~\cite{sparsevlm}; and the last one is the proposed \textit{dynamic prior}, which is recomputed at every pruning stage adaptively. Among all baselines, our dynamic prior reaches the best performance on all four benchmarks.

We further analyze the pruning schedule. Besides the default setting with \(K=3\), we compare variants with fewer or more pruning stages (\(K=2\) and \(K=4\)) and perturb each pruning layer around the selected schedule. Table~\ref{tab:ablation2} shows that the observation-driven schedule \((1,10,20)\) achieves strong performance without exhaustive hyperparameter search. While nearby variants can occasionally obtain comparable scores on individual benchmarks, the selected schedule consistently lies in a high-performing region, supporting our claim that attention-shift analysis provides a simple and effective way to identify pruning locations.

\section{Conclusion}
This paper proposes AdaptInfer, a plug-and-play solution for accelerating VLM inference via dynamic text-guided token pruning. Moreover, we observe a highly consistent pattern of cross-modal attention shifts in VLMs and introduce a data-driven heuristic pruning schedule accordingly. AdaptInfer introduces minimal additional computational overhead while maintaining higher accuracy than existing approaches. For example, AdaptInfer reduces CUDA latency by 61.3\% while retaining an average accuracy of 91.5\% on LLaVA-1.5-7B. On Qwen2-VL-2B, it retains an average accuracy of 99.4\% even when only 30\% of the vision tokens are retained on average across layers.

\section*{Limitations}

\paragraph{Attention Accessibility.}
AdaptInfer relies on accessing selected attention blocks during inference, including t2t and t2v attentions. Therefore, AdaptInfer can be directly applicable to open-source VLMs or deployment frameworks in principle (we have evaluated its generalization on LLaVA-1.5, Qwen2-VL, and InternVL3), but it is hard to be applicable to closed-source or API-only systems.

\paragraph{Pruning Schedule Transferability.}
Although AdaptInfer avoids expensive grid search for pruning locations, its pruning schedule is still derived from an offline attention-shift analysis. As a result, applying the method to a new architecture may also require a lightweight calibration step.

\paragraph{Aggressive Pruning.}
AdaptInfer improves the accuracy-efficiency trade-off, but aggressive pruning can still remove fine-grained visual evidence. This can affect tasks that rely on dense or localized visual information, such as counting and small-object recognition. Moreover, multi-turn dialogue may present an extra challenge: visual regions that are irrelevant to the current query could become important in subsequent turns, while their corresponding tokens may have already been removed.

\paragraph{Attention as Importance Proxy.}
Our method uses attention scores as an efficient proxy for token importance. While this proxy works well empirically, attention weights may not always reflect the causal contribution of each token to the final prediction, which remains a broader limitation of attention-based pruning methods.

\section*{Ethical Considerations}

This work aims to advance the grounding of LLMs in vision. We do not introduce new datasets, collect human-subject data, or use private or sensitive data. This work may make the multimodal LLMs easier and cheaper to deploy. Our method builds on existing VLMs and does not introduce additional ethical risks beyond those already associated with the underlying models.

We use publicly available models, datasets, and baseline implementations in accordance with their respective licenses and terms of use.

\section*{Acknowledgment}
This work was supported in part by the National Science and Technology Major Project under Grant 2022ZD0114903, and by National Natural Science Foundation of China under Grants No. 62432008 and 62472248.

The authors declare that they use generative AI tools for grammar refinement and language polishing only. The authors take full responsibility for the entire content of this paper.

\bibliography{custom}

\appendix
\section{Appendix Statement}
The following appendix sections provide supplementary technical materials intended to support reproducibility, including implementation details, experimental configurations, and auxiliary analyses not essential to the main claims.

\section{Change-Point Detection}
\label{appendix:cpd}

To identify shifts in cross-attention across transformer layers, we employ change-point detection~\citep{changpoint} on cumulative cross-attention curves. The intuition is that a significant change in attention behavior corresponds to a noticeable inflection in the cumulative attention distribution over layers.

Given a token's cross-attention curve across $L$ layers, denoted as:
\begin{equation}
  \mathbf{a} = [a_1, a_2, \dots, a_L], \quad \text{where } a_\ell \in \mathbb{R}  
\end{equation}

we first compute the cumulative attention:
\begin{equation}
  \mathbf{c} = \left[ \sum_{i=1}^1 a_i, \sum_{i=1}^2 a_i, \dots, \sum_{i=1}^L a_i \right]  
\end{equation}

This sequence $\mathbf{c}$ captures the aggregate build-up of attention, which we treat as a univariate time series. We then apply change-point detection to identify the most likely layer index where a shift in trend occurs.

We adopt the \texttt{ruptures} Python library\footnote{\url{https://github.com/deepcharles/ruptures}} for offline detection. Specifically, we use the Binary Segmentation (Binseg) algorithm with an $\ell_2$ cost model. This algorithm aims to find a segmentation point that minimizes the within-segment variance. Formally, given the cumulative curve $\mathbf{c} = [y_1, y_2, \dots, y_L]$, we solve:

\begin{equation}
\min_{b} \left( \sum_{t=1}^{b} (y_t - \mu_1)^2 + \sum_{t=b+1}^{L} (y_t - \mu_2)^2 \right),
\end{equation}

where $\mu_1$ and $\mu_2$ are the mean values of the first and second segments, respectively. This corresponds to the $\ell_2$ cost model used in ruptures, which measures the homogeneity of each segment by its squared deviation from the mean.

We restrict the number of change points to 1, making the detection both efficient and robust. The returned breakpoint $b \in \{1, \dots, L\}$ indicates the first significant shift in attention accumulation for the given token. This layer index is treated as the token’s \textit{cross-attention shift point}, which guides our downstream pruning strategy.

\section{Analysis of Computational Complexity}\label{sec:compute}
Assuming \(n=T+V\) denotes the current sequence length, \(d\) denotes the VLM's hidden state dimension, and \(m\) denotes the hidden size of projection layer in the FFN network. For each transformer layer in the prefill stage, the FLOPs can be estimated by
\begin{equation}
    {FLOPs^{\text{prefill}}} = 4nd^2+2n^2d+3ndm.
\end{equation}

For each pruning layer, the additional FLOPs of are computed below:
\begin{equation}
    {FLOPs^{\text{prune}}} = T^2 + 2TV.
\end{equation}
Then, during the decode stage, the FLOPs of each layer can be estimated by
\begin{equation}
    {FLOPs^{\text{decode}}} = 4d^2+2nd+3dm.
\end{equation}

From the above analysis, the computational bottleneck of VLM inference primarily arises in the prefill stage, where the self-attention operation incurs a quadratic complexity term \(O(n^2 d)\). 
Since the introduction of a large number of vision tokens substantially increases the sequence length \(n\), the prefill stage becomes significantly more expensive than the decode stage, whose complexity scales only linearly with \(n\). Therefore, to effectively reduce overall inference cost, vision token pruning should be performed during the prefill stage rather than deferred to the decode stage.

\section{Compatibility with FlashAttention}\label{sec:flashattention}

Many modern VLMs rely on FlashAttention~\citep{dao2022flashattention} for efficient prefill and therefore do not explicitly materialize attention matrices. Since AdaptInfer uses attention for token ranking, a natural concern is whether this would introduce substantial overhead and offset the inference gains brought by pruning. Our answer is negative. AdaptInfer does \emph{not} require attention extraction from all transformer layers, nor does it require the full attention matrix at any queried layer. In practice, attention is queried only at the selected pruning layers (at most three in our experiments), and only the \textit{text-to-text} ($t2t$) and \textit{text-to-vision} ($t2v$) blocks are needed. The expensive \textit{vision-to-vision} ($v2v$) block is never used.

We build on the dual-pass FlashAttention strategy introduced by SparseVLM~\citep{sparsevlm}, but adapt it to our dynamic text-guidance setting. The first pass is the standard FlashAttention forward pass, which produces the hidden states used by the model. At a queried pruning layer, we then perform an auxiliary pass that reuses the same queries and keys, but replaces the value matrix with task-specific selector matrices so that only the required attention sub-blocks are returned. This preserves FlashAttention compatibility while avoiding explicit materialization of the full attention map.

\vspace{0.3em}
\noindent\textbf{Required attention blocks.}
Let the input sequence at a queried layer contain $T$ text tokens and $V$ vision tokens, with total length $L=T+V$. For head $h$, let
\begin{equation}
A^{(h)}=\mathrm{Softmax}\!\left(\frac{Q^{(h)}K^{(h)\top}}{\sqrt{d}}\right)\in\mathbb{R}^{L\times L}
\end{equation}
denote the self-attention matrix, and let
\begin{equation}
\bar{A}=\frac{1}{H}\sum_{h=1}^{H}A^{(h)}
\end{equation}
be the attention matrix averaged over heads. Since the FlashAttention-compatible dual-pass returns mean attention statistics, AdaptInfer operates on $\bar{A}$ in this appendix.

Let $S_q\in\{0,1\}^{T\times L}$ select the rows corresponding to text queries, let $S_t\in\{0,1\}^{L\times T}$ select text-key columns, and let $S_v\in\{0,1\}^{L\times V}$ select vision-key columns. Then the only two attention blocks required by AdaptInfer are
\begin{equation}
\bar{A}_{t2t}=S_q\bar{A}S_t\in\mathbb{R}^{T\times T},
\end{equation}
\begin{equation}
   \bar{A}_{t2v}=S_q\bar{A}S_v\in\mathbb{R}^{T\times V} 
\end{equation}

The dynamic text prior is computed as
\begin{equation}
w=\mathbf{1}^{\top}\bar{A}_{t2t}\in\mathbb{R}^{T},
\end{equation}
and the visual-token importance scores are then
\begin{equation}
s=w^{\top}\bar{A}_{t2v}\in\mathbb{R}^{V}.
\label{eq:flash_s}
\end{equation}
Equivalently,
\begin{equation}
s=\left(\sum_{i=1}^{T}\bar{A}_{t2t}[i,:]\right)^{\top}\bar{A}_{t2v}.
\end{equation}
This is the mean-attention version of Eq.~\ref{eqa:2} and Eq.~\ref{eqa:3} in the main paper.

\vspace{0.3em}
\noindent\textbf{Dual-pass extraction of the required blocks.}
In the auxiliary pass, we keep the same queries and keys as in the original layer, but replace the value matrix by selector matrices. Specifically, if we set the value matrix to
\begin{equation}
V_t=S_t,
\end{equation}
then the FlashAttention output restricted to text-query rows becomes
\begin{equation}
O_t=S_q\bar{A}V_t=S_q\bar{A}S_t=\bar{A}_{t2t}.
\end{equation}
Likewise, if we set the value matrix to
\begin{equation}
V_v=S_v,
\end{equation}
the output restricted to text-query rows is
\begin{equation}
O_v=S_q\bar{A}V_v=S_q\bar{A}S_v=\bar{A}_{t2v}.
\end{equation}
Therefore, the two auxiliary queries recover exactly the two sub-blocks required by AdaptInfer, without constructing the full attention matrix. In other words, our modification of the SparseVLM dual-pass is to query \emph{block-restricted mean attention} tailored to dynamic text guidance, rather than extracting generic attention statistics for all tokens.

\vspace{0.3em}
\noindent\textbf{Complexity.}
A full attention matrix at one layer has size
\begin{equation}
L^2=(T+V)^2=T^2+2TV+V^2,
\end{equation}
which includes the expensive $v2v$ block of size $V^2$. By contrast, AdaptInfer only needs the two blocks $\bar{A}_{t2t}$ and $\bar{A}_{t2v}$, whose total size is
\begin{equation}
T^2+TV.
\end{equation}
Hence, the auxiliary extraction cost per queried layer scales with the required block size
\begin{equation}
O(T^2+TV),
\end{equation}
rather than
\begin{equation}
O((T+V)^2)=O(T^2+2TV+V^2).
\end{equation}
Since in VLM inference we typically have $T\ll V$, avoiding the $V^2$ term is crucial, especially for high-resolution images and video inputs. Moreover, this auxiliary extraction is performed only at a small number of pruning layers, instead of every transformer layer. If the number of queried pruning layers is $M$, the total auxiliary extraction cost is
\begin{equation}
O\!\left(M(T^2+TV)\right),
\end{equation}
with $M\le 3$ in all our experiments.

In summary, AdaptInfer remains compatible with FlashAttention by extending the dual-pass design of SparseVLM to the specific needs of dynamic text guidance. The method only queries the $t2t$ and $t2v$ blocks at a few selected layers and never uses the dominant $v2v$ block. As a result, the additional overhead is modest in practice and does not materially offset the efficiency gains from pruning.

\section{Experiments on LLaVA-1.5-13B}\label{13b}
\subsection{Attention Shift Detection on LLaVA-1.5-13B}

\begin{figure}[t]
    \centering
    \includegraphics[width=0.9\linewidth]{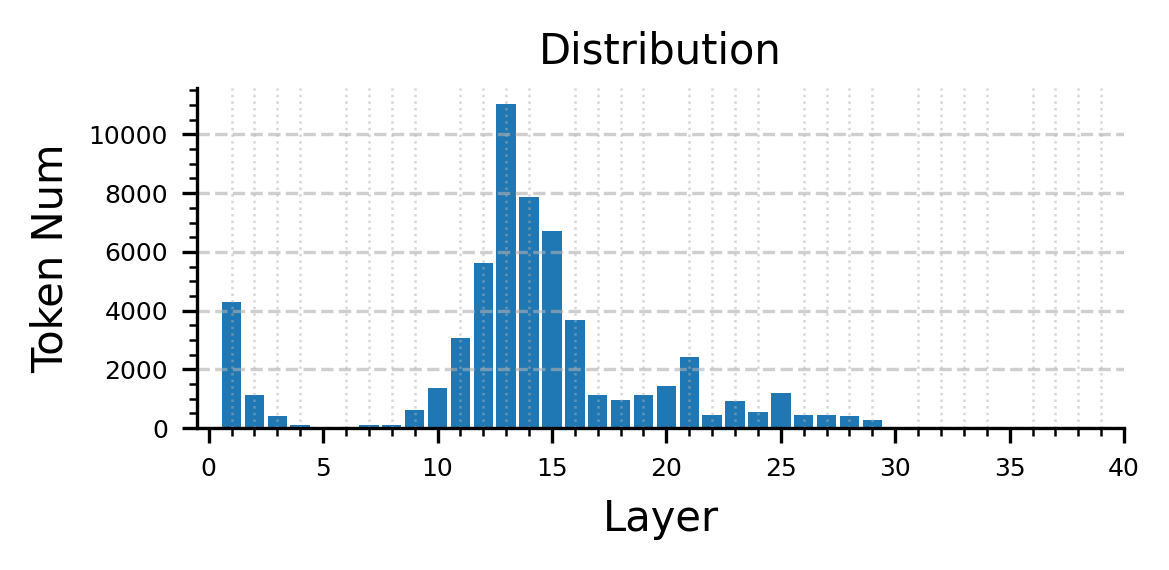}\\
    \small (a) Attention shift distribution on MME.\\

    \includegraphics[width=0.9\linewidth]{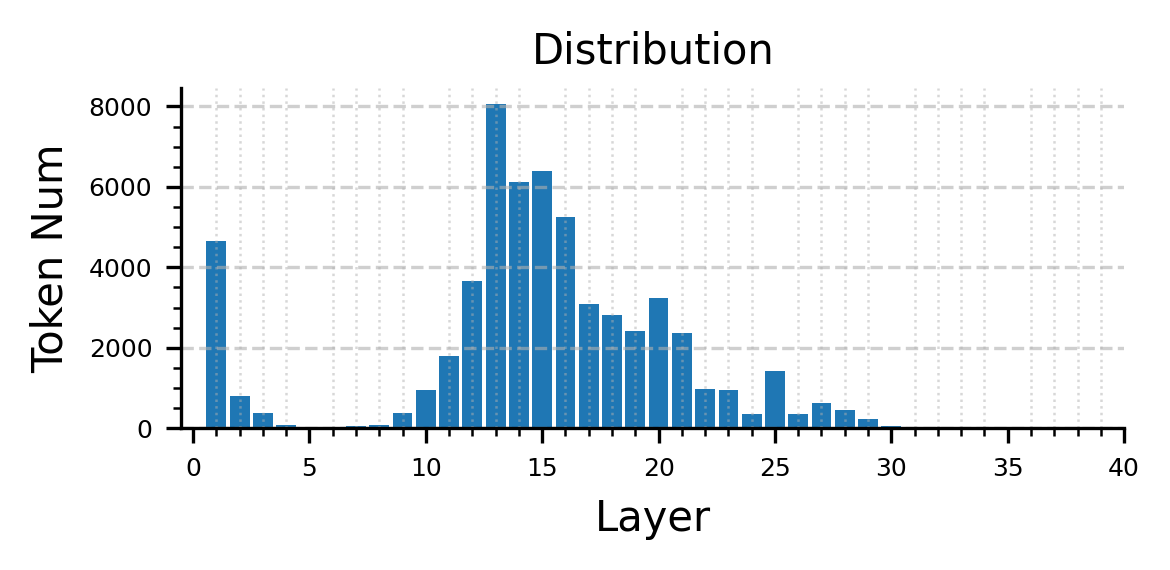}\\
    \small (b) Attention shift distribution on TextVQA.
    
    \caption{\textbf{Layer-wise distribution of attention shifts on LLava-1.5-13B.}}
    \label{fig:shift_appendix}

\end{figure}

To validate the transferability of our proposed pruning schedule, we conduct attention shift detection on LLaVA-1.5-13B~\citep{liu2023llava}, which is built upon a 40-layer LLaMA-2-13B backbone~\citep{touvron2023llama2openfoundation}. The resulting distributions on two benchmark datasets, MME and TextVQA, are presented in Figure~\ref{fig:shift_appendix}. We observe that the distribution of attention shift points still remains highly consistent on MME and TextVQA. The early-stage concentration of attention shifts consistently occurs at layer 1. Meanwhile, in later layers, dense change points are slightly delayed compared to those in LLaVA-1.5-7B, which typically occurs around layers 10 to 22. This suggests that for LLaVA-1.5-13B and other VLMs built on the LLaMA-2-13B backbone, an appropriate pruning schedule should select representative layers such as 1, 11, and 22 accoring to our proposed schdule.

Moreover, our attention shift detection pipeline is both simple and lightweight. For example, on the MME dataset, extracting 1000 attention maps and performing change-point detection takes only about 8 minutes in total. This confirms the efficiency and transferability of our pruning schedule across different model scales and architectures.

\subsection{Performance of AdaptInfer on LLava-1.5-13B}\label{13b-exp}
Using the parameters above, we test our AdaptInfer on LLava-1.5-13B to further validate the effectiveness of the proposed pruning schedule. The experiments are based on three widely used dataset MME~\citep{mme}, MMBench (MMB)~\citep{mmbench}, and TextVQA~\citep{textvqa} under 128 and 64 vision token budgets. The results are shown in Table~\ref{tab: llava13b}.

\begin{table}[t]
\small
\centering
\begin{tabular}{l|ccc|c}
\toprule
\textbf{Tokens}  & \textbf{MME} &\textbf{MMB}&\textbf{TextVQA}& \textbf{Ratio (\%)} \\
\midrule
 576        &1826& 61.2 & 68.5 & 100 \\
\midrule
 128 & 1813 & 59.8 & 67.4 & 98.5\\

64 & 1735& 58.1 & 66.2  & 95.5\\
\bottomrule
\end{tabular}
\caption{\textbf{Performance of AdaptInfer on LLava-1.5-13B.} Our method still maintain high overall accuracy.}
\label{tab: llava13b}
\end{table}

Under the more demanding 13-billion-parameter setting of LLaVA-1.5, AdaptInfer continues to deliver high performance. As Table~\ref{tab: llava13b} shows, pruning the vision token numbers from 576 to 128 on average per layer results in only a 1.5\% drop in overall accuracy (from 100\% to 98.5\%). Even with an aggressive 64 vision token budget, AdaptInfer achieves an 8.9× reduction in vision-token count, the model still retains 95.5\% of its full-token accuracy. These numbers demonstrate that our token-adaptive pruning solution scales gracefully with model size, confirming AdaptInfer’s robustness.

\section{Attention Shift Detection on Qwen2-VL-2B}\label{qwen2b}
\begin{figure}[ht]
    \centering
    \includegraphics[width=0.9\linewidth]{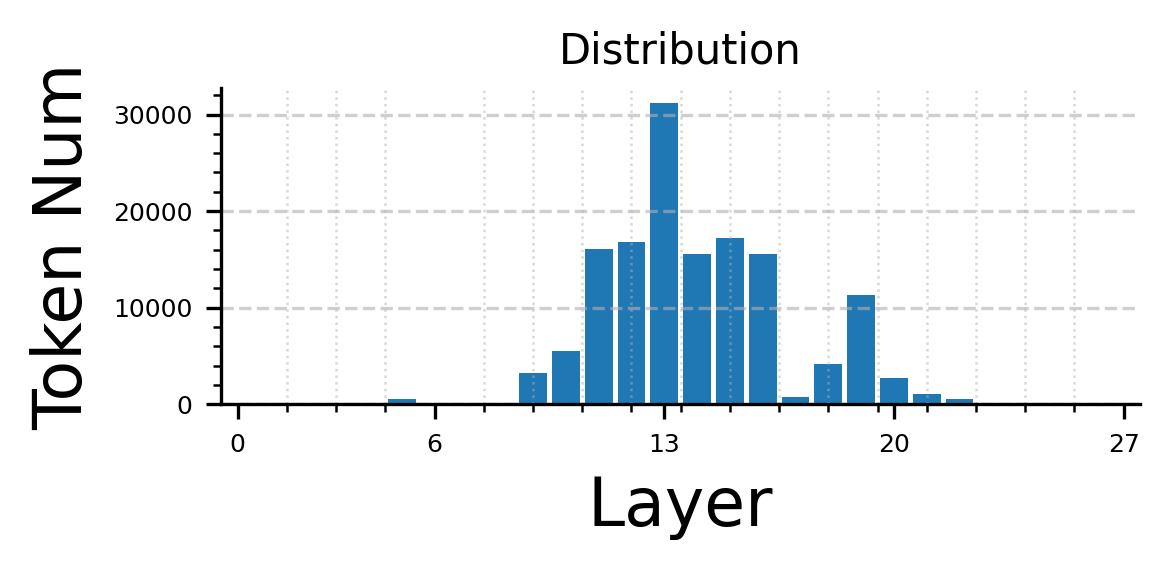}
    \caption{\textbf{Layer-wise distribution of attention shifts on Qwen2-VL-2B on MME.}}
    \label{fig:appendix_qwen}
\end{figure}

We also conduct our attention shift detection on a different VLM framework, Qwen2-VL-2B~\citep{Qwen2-VL}, which contains 28 transformer layers in its LLM backbone. The attention shift distribution on MME is shown in Fig~\ref{fig:appendix_qwen}. The high-frequency regions are among layer 11-19, where stable regions are among layer 0-10 and 20-28.
Based on our proposed pruning schedule, we select the token pruning locations at layer 0, 9, and 19. Moreover, conducting the attention shift analysis on Qwen2-VL- 2B takes only about 5 minutes in total, which illustrates the lightweightness and transferability of the proposed attention shift analysis.

\section{Pruning Tolerance}
To further understand the vision token pruning, we conduct a pruning-tolerance evaluation. Specifically, we apply AdaptInfer to Qwen2-VL-2B and report the performance across 14 multimodal tasks in the MME benchmark, comparing the scores with and without pruning. The results presented in Table~\ref{appendix:tab_tol} indicate that code reasoning, count and color-based tasks have a lower pruning tolerance relatively. Meanwhile, calculation and OCR tasks outperforms the original Qwen due to higher pruning tolerance and the noise reduction effect by token pruning.

\begin{table}[ht]
\small
\centering
\caption{\textbf{Pruning-tolerance evaluation on MME.} For AdaptInfer, 30\% tokens are retained averagely.}
\label{appendix:tab_tol}
\begin{tabular}{lccc}
\toprule
\textbf{Task} & \textbf{Vanilla} & \textbf{AdaptInfer} & \textbf{Ratio} \\
\midrule
Existence      & 200 & 200 & 100\% \\
Count          & 145 & 135 & 93.1\% \\
Position       & 158 & 158 & 100\% \\
Color          & 190 & 180 & 94.7\% \\
Posters        & 125 & 123 & 98.4\% \\
Celebrity      & 138 & 136 & 98.6\% \\
Scene          & 159 & 159 & 100\% \\
Landmark       & 174 & 172 & 98.9\% \\
Artwork        & 135 & 135 & 100\% \\
OCR            & 88  & 103 & 117\% \\
Commonsense    & 112 & 111 & 99.1\% \\
Calculation    & 23  & 33  & 143\% \\
Translation    & 163 & 155 & 95.1\% \\
Code Reasoning & 93  & 85  & 91.4\% \\
\bottomrule
\end{tabular}
\end{table}

\section{Instance-Level Decision Analysis}

There is an inherent trade-off between efficiency and performance in vision token pruning. we admit that aggressive token pruning can induce instance-level decision changes. To examine this effect, we perform a simple instance-level analysis comparing the vanilla Qwen2-VL-2B and 30\% token retained AdaptInfer on MME. For each example, we record whether the prediction is (correct/error) before and after pruning. We tabulate the transition, and report the change ratios in Table~\ref{tab:appendix_ilchange}. We find that the vast majority of predictions remain unchanged, while the fraction of Correct to Error cases is small and comparable to the number of Error to Correct cases. 

\begin{table}[ht]
\small
\centering
\caption{\textbf{Instance-Level change before and after pruning.} C: Correct; E: Error.}
\label{tab:appendix_ilchange}
\begin{tabular}{lcccc}
\toprule
 & \textbf{C to C} & \textbf{C to E} & \textbf{E to C} & \textbf{E to E} \\
\midrule
\textbf{Percentage} & 78.2\% & 1.7\% & 1.3\% & 18.8\% \\
\bottomrule
\end{tabular}
\end{table}

\section{Visualization}

\begin{figure*}[ht]
    \centering
    \includegraphics[width=0.9\linewidth]{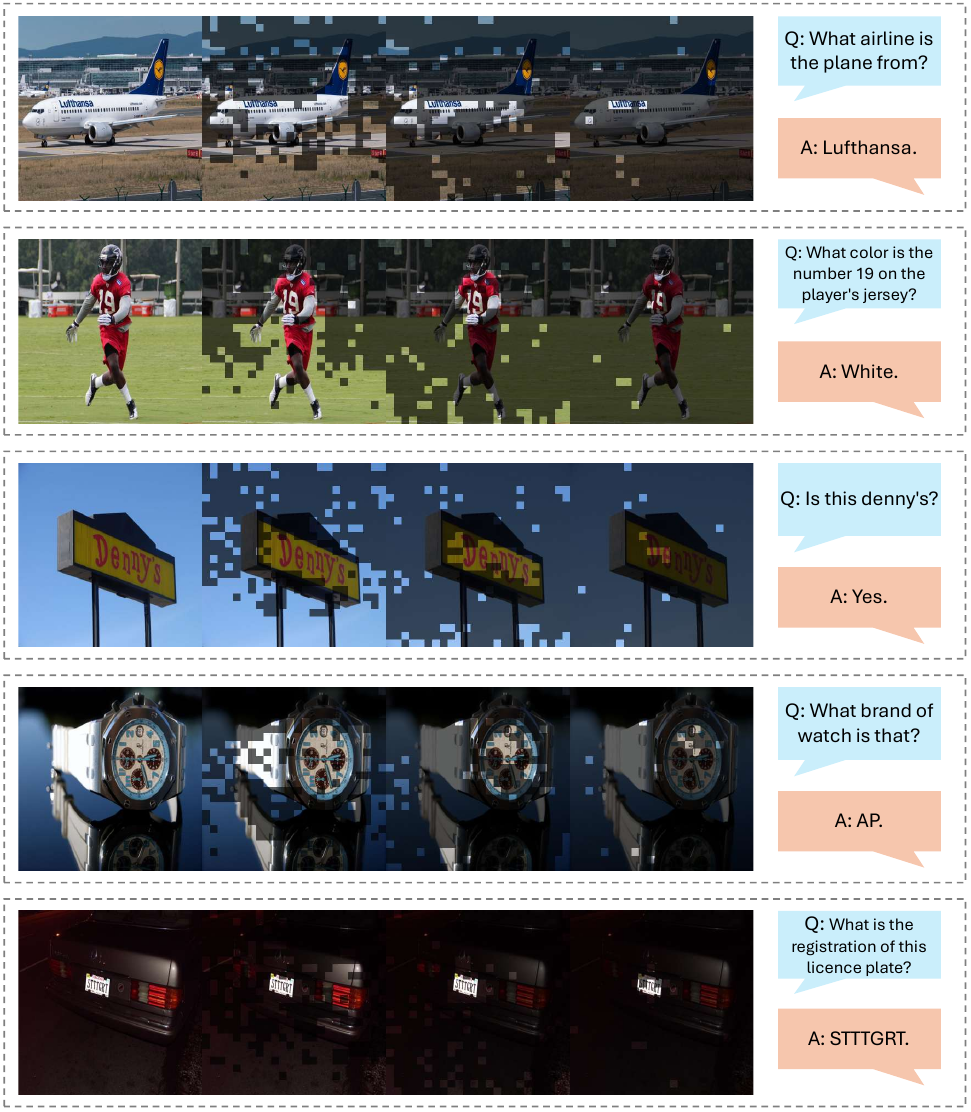}
    \caption{\textbf{Visualization of AdaptInfer on different examples.}}
    \label{fig:appendix_vis}
\end{figure*}

To qualitatively assess the pruning effect of the proposed AdaptInfer, we conduct an additional visualization study. In line with the main paper, vision tokens are pruned after the 1st, 10th, and 20th transformer layers. The resulting figures depict the spatial distribution of the remaining tokens at each stage: transparent regions correspond to tokens that survive pruning, whereas semi-transparent patches indicate positions that have been discarded.

The experiment is carried out on the TextVQA~\citep{textvqa} benchmark, from which we sample a representative set of examples for display. As the layers deepen, \textsc{AdaptInfer} rapidly suppresses tokens associated with semantically irrelevant background areas while consistently preserving those aligned with question-critical cues. The visualization is shown in Figure~\ref{fig:appendix_vis}.

\section{Performance Analysis}
\begin{figure*}[ht]
    \centering
    \includegraphics[width=1\linewidth]{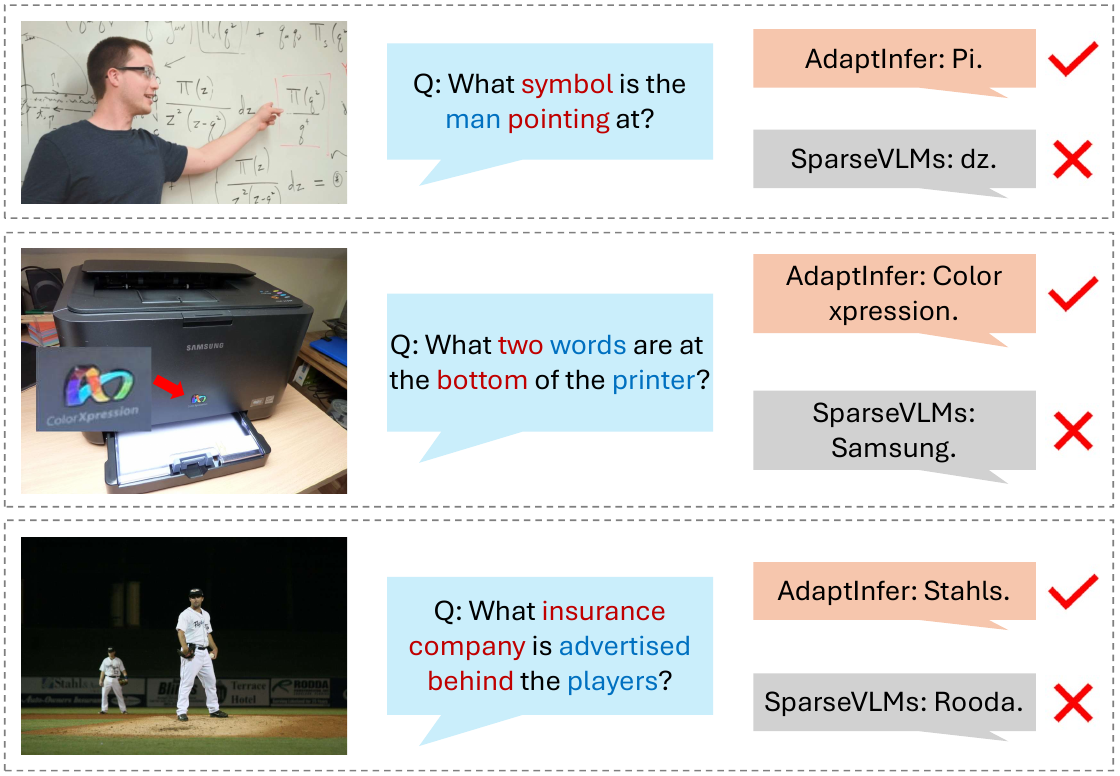}
    \caption{\textbf{Examples where AdaptInfer outperforms SparseVLM.} AdaptInfer performs better when text prompts are rich in descriptive modifiers.}
    \label{fig:appendix_vis2}
\end{figure*}

We provide further analysis of the advantages of our dynamic text guidance method, AdaptInfer, compared with the static text guidance solution, SparseVLM~\citep{sparsevlm}. We extract some VQA examples from the TextVQA dataset~\citep{textvqa}, and the answers of both methods are presented in Figure~\ref{fig:appendix_vis2}.  

Within each question, blue highlights denote the static key text tokens selected offline by SparseVLM, whereas red highlights trace the key text tokens adaptively chosen from various layers by AdaptInfer. Close inspection reveals a consistent pattern: SparseVLM almost exclusively locks onto the text tokens that bear the strongest direct semantic link to the visual scene (e.g., \textit{printer} or \textit{players}), but often discards adjectival or adverbial modifiers that disambiguate entities (e.g., \textit{pointing}, \textit{two} or \textit{behind}). When the prompt becomes syntactically richer and contains multiple content nouns and stacked modifiers static text guidance leads to an under-specified textual context and, consequently, a notable drop in answer accuracy. Conversely, AdaptInfer updates its key text token set dynamically during inference, rarely missing any prominent guidance. This dynamical adjustment make AdaptInfer overcome such limitation.

\section{Experiment Setup}
In this appendix section, we will briefly introduce the details of our experiment setups.

\subsection{Hardware Environment}
To facilitate reproducibility, we ran every experiment on a single workstation that hosts an Intel Xeon Platinum 8358P processor together with eight NVIDIA RTX 4090 GPUs, each furnished with 24 GB of VRAM. The system operates under Ubuntu Linux 24.04, and all code was compiled using GCC 13.2.0 with Binutils 2.42 as the linker. 



\subsection{General Implementation Details}
In our evaluation of AdaptInfer, the pruning ratios at each pruning location are the hyperparameters need to be selected as well. For these hyperparameters, we follow the same rank-based sparsification-level adaptation method which SparseVLM~\citep{sparsevlm} introduced.

In the paper, we evaluated four vision–language models in total, including LLaVA-1.5-7B, LLaVA-1.5-13B, InternVL3-2B~\citep{internvl3} and Qwen2-VL-2B. To stay within GPU-memory limits, we ran the LLava-7B variant, Qwen2-VL-2B, and InternVL3-2B on one GPU, and the LLava-13B variant on three GPUs. All training and inference are performed in torch.float16 or torch.bfloat16 precision to further conserve memory usage. All the results of AdaptInfer reported in the context are the average of five runs.

\subsection{Experiments on DART}\label{dart}
DART~\citep{dart} reports performance on LLaVA-v1.5-7B under different retained token counts in the paper. However, these values refer to the number of remaining vision tokens after pruning, whereas our work defines the token budget as the average number of vision tokens retained per layer. Accordingly, our token budget measures the mean token preservation across all transformer layers in the prefill stage rather than the final post-pruning token count.

To enable a fair and consistent comparison, we re-implemented DART using the official codebase while keeping all default hyperparameters unchanged, including pruning location $K$ is set to 2, $pivot\_image\_token$ is set to 4, and $pivot\_text\_token$ is set to 4. We only change the $max\_num\_trunction$ to 98 for a token budget of 128 and to 26 for a token budget of 64 respectively, ensuring that the resulting average retained token numbers match our budget definition.

\subsection{Implementation Details of AdaptInfer on Qwen2-VL-2B}
\label{appendix:qwen_detail}
Qwen2-VL-2B employs a dynamic number of vision tokens, ranging from 4 to 16,384 per image. To ensure stable performance and prevent GPU memory overflow, we constrain the vision token count within a bounded range, setting the minimum to 256 and the maximum to 1,280. For video-based multimodal evaluation, we follow FastV~\citep{fastv} and SparseVLM~\citep{sparsevlm} and use the first 1,000 test samples from each benchmark to reduce evaluation time. The frame rate for all videos is fixed at 4 FPS.

\end{document}